# Wearable Vision Detection of Environmental Fall Risks using Convolutional Neural Networks

Mina Nouredanesh, Andrew McCormick, Sunil L. Kukreja, *Senior Member*, and James Tung, *Member*

*Abstract*—In this paper, a method to detect environmental hazards related to a fall risk using a mobile vision system is proposed. First-person perspective videos are proposed to provide objective evidence on cause and circumstances of perturbed balance during activities of daily living, targeted to seniors. A classification problem was defined with 12 total classes of potential fall risks, including slope changes (e.g., stairs, curbs, ramps) and surfaces (e.g., gravel, grass, concrete). Data was collected using a chest-mounted GoPro camera. We developed a convolutional neural network for automatic feature extraction, reduction, and classification of frames. Initial results, with a mean square error of 8%, are promising.

## I. INTRODUCTION

A number of hazards in the home and public environment have been identified that contribute to falls and related injuries. These influences interact with intrinsic factors, such as poor vision or balance, to compound fall risk for seniors. A major challenge is a lack of available tools to assess individual risk, particularly the frequency of exposure to specific hazards. For example, an individual with unstable gait may be at greater risk of falls with frequent exposure to hazards than another with similar capabilities but little or no exposure. The aim of the current study is to examine the potential of wearable egocentric cameras (e.g., GoPro), coupled with advances in machine learning techniques, to detect fall risk hazards.

## II. MATERIALS AND METHODS

Healthy young participants were asked to walk around the University of Waterloo campus using a GoPro Hero4 camera attached on participants' chest (Fig. 1). Test data comprised of 3669 video frames, and annotated with the following labels: crosswalk, curbs, ramp, stairs (ascending), stairs (descending), gravel, concrete, tiles, bricks, carpets, snow, and rocks, as a 12-class problem.

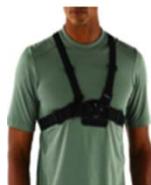

Figure 1- GoPro Hero4 camera attached to on chest.

### A. Convolutional neural networks for detection task

To detect which environmental factor is presented in a given frame, we explore the use of Convolutional Neural Networks (CNN), a class of biologically-inspired variants of multilayer perceptrons (MLPs) that resemble the human visual cortex. After converting RGB images to grayscale and resizing the frames into 32x32 images, we developed a pyramidal network [1] with 3 convolutional layers (C1,C3,C5), 2 sub-sampling (max-pooling) layers (S2,S4), followed by a fully connected MLP (F6) with 12 neurons in output (Figure 2). We employed 64, 32, and 16 layers of a size 5x5, 3x3, and 6x6 for the convolution layers C1, C3, C5 respectively. Connections between each feature map in convolution layer and its adjacent to sub-sampling layer is 1-to-1. The final layer F6 is fully connected to C5. Activation functions for network layers are set as tangent sigmoid (tansig) for C1, C3, and C5 convolution layers, linear transfer function (purelin) for S2 and S4 subsampling layers, and "tansig" for output layer. Resilient backpropagation training method was chosen as a fast first-order training algorithm.

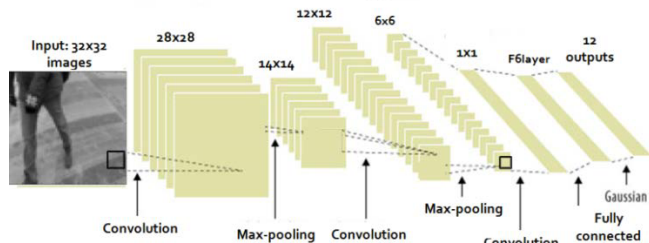

Figure 2- Topology of the developed convolutional neural network

## III. RESULTS

During the process of training in 100 epochs, mean square error (MSE) converges to ~0.08, indicating accuracy of the trained network in detection of objects/terrains, roughly interpreted as a mean classification accuracy of 92%.

## IV. DISCUSSION

This study examined the potential for wearable egocentric cameras, combined with machine learning techniques, to automatically detect fall risk hazards. Previous methods have relied manual identification of environmental circumstances, which is inefficient and impractical for real-time use [2]. To the best of the authors' knowledge, this study is the first to develop an automated method for fall hazard detection with promising preliminary results.

All training was conducted using grey-scale images which significantly reduced processing time but likely reduced the potential to distinguish risks sharing similar texture, shape or color (e.g. bricks and tiles, or crosswalk and stairs). Next steps will examine the use of CNN on RGB images.

Research supported by National Sciences and Engineering Research Council of Canada (NSERC) and the Network for Aging Research (University of Waterloo). Mina Nouredanesh, Andrew McCormick, and James Tung are with the Department of Mechanical and Mechatronics Engineering, University of Waterloo, email: james.tung@uwaterloo.ca)

Sunil Kukreja is the Head of Neuromorphic Engineering and Robotics, Singapore Institute for Neurotechnology (SINAPSE), National University of Singapore (NUS), Singapore, (email: sunilkukreja.sinapse@gmail.com)